\documentclass[lettersize,journal]{IEEEtran}
\usepackage{amsmath,amsfonts}
\usepackage[linesnumbered]{algorithm2e}
\usepackage{array}
\usepackage[caption=false,font=normalsize,labelfont=sf,textfont=sf]{subfig}
\usepackage{textcomp}
\usepackage{stfloats}
\usepackage{url}
\usepackage{hyperref} 
\usepackage{verbatim}
\usepackage{graphicx}
\usepackage{cite}
\usepackage{color}

\begin{document}

\title{PD-SORT: Occlusion-Robust Multi-Object Tracking Using Pseudo-Depth Cues}

\author{Yanchao Wang, Dawei Zhang, Run Li, Zhonglong Zheng,~\IEEEmembership{Member,~IEEE}, Minglu Li,~\IEEEmembership{Fellow,~IEEE}
\thanks{Yanchao Wang, Dawei Zhang, Run Li, Zhonglong Zheng and Minglu Li are with the School of Computer Science and Technology, Zhejiang Normal University, Jinhua 321004, China. The corresponding author is Dawei Zhang (Email: davidzhang@zjnu.edu.cn).}
}

\markboth{Journal of \LaTeX\ Class Files,~Vol.~14, No.~8, August~2021}%
{Shell \MakeLowercase{\textit{et al.}}: A Sample Article Using IEEEtran.cls for IEEE Journals}


\maketitle

\begin{abstract}
Multi-object tracking (MOT) is a rising topic in video processing technologies and has important application value in consumer electronics. Currently, tracking-by-detection (TBD) is the dominant paradigm for MOT, which performs target detection and association frame by frame.
However, the association performance of TBD methods degrades in complex scenes with heavy occlusions, which hinders the application of such methods in real-world scenarios.To this end, we incorporate pseudo-depth cues to enhance the association performance and propose Pseudo-Depth SORT (PD-SORT). First, we extend the Kalman filter state vector with pseudo-depth states. Second, we introduce a novel depth volume IoU (DVIoU) by combining the conventional 2D IoU with pseudo-depth. Furthermore, we develop a quantized pseudo-depth measurement (QPDM) strategy for more robust data association. Besides, we also integrate camera motion compensation (CMC) to handle dynamic camera situations. With the above designs, PD-SORT significantly alleviates the occlusion-induced ambiguous associations and achieves leading performances on DanceTrack, MOT17, and MOT20. Note that the improvement is especially obvious on DanceTrack, where objects show complex motions, similar appearances, and frequent occlusions. 
The code is available at \href{https://github.com/Wangyc2000/PD_SORT}{https://github.com/Wangyc2000/PD\_SORT}.
\end{abstract}


\section{Introduction}
\IEEEPARstart{M}{ulti-Object} tracking (MOT) aims to detect all desired objects in a video and maintain their identities across frames, which serves as a fundamental vision task. With the rapid development of consumer technologies, MOT systems can be deployed to diverse edge devices with cameras (e.g. smartphones, automobiles, drones, etc.), enabling vast applications for consumer electronics including but not limited to autonomous driving \cite{luo2021exploring}, video surveillance \cite{kim2012multi,iepure2021novel}, UAV applications \cite{yang2023bandt}, and human behavior analysis \cite{zhao2023human}.
Nevertheless, complex object motions and dense crowds still pose challenges for the real-world application of MOT methods.

\begin{figure*}
    \centering
    \includegraphics[width=0.9\textwidth]{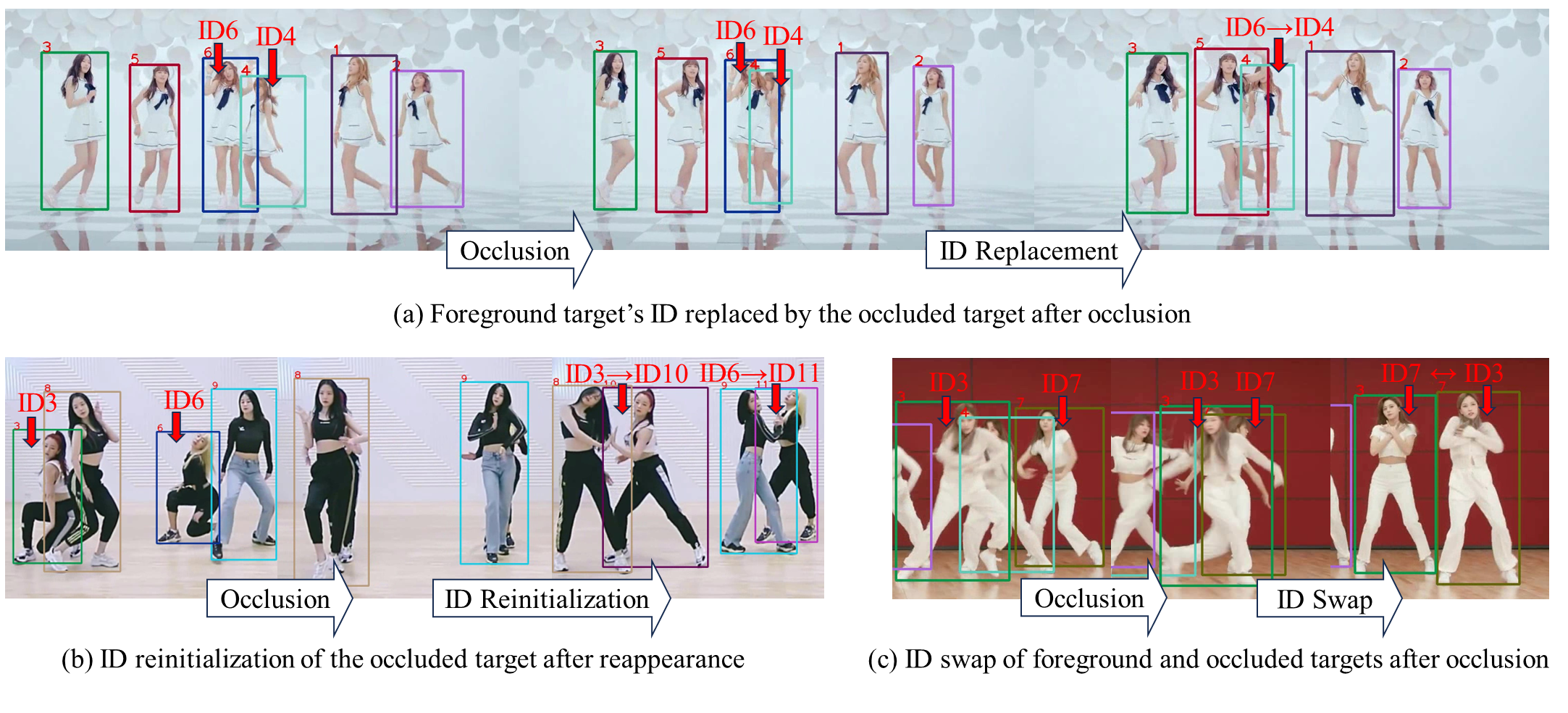}
    \caption{Three examples of occlusion-induced tracking failures. The samples are OC-SORT results on DanceTrack, where objects have diverse motions and similar appearances.  }
    \label{fig:occlusion-induced problems}
\end{figure*}

Currently, tracking-by-detection (TBD) \cite{bewley2016simple,bochinski2017high,wojke2017simple,zhang2022bytetrack,cao2023observation} is the dominant paradigm for solving the MOT problem. 
Methods following the TBD paradigm decompose tracking into two sub-steps: i) performing frame-by-frame object detection, and ii) matching the detected objects across frames using association algorithms to form trajectories. Typically, the detection task is realized using off-the-shelf object detectors \cite{ren2015faster,ge2021yolox}, and the association task is achieved by bipartite graph matching with the Hungarian algorithm \cite{bipmatch}, where motion cues and appearance cues are used for similarity evaluation.
However, in complex scenarios with crowded objects and non-linear motion (e.g., scenes from the DanceTrack \cite{sun2022dancetrack} benchmark), occlusions happen frequently. In such cases, bounding boxes of intersecting objects in 2D images are highly overlapped, motion models in TBD methods based on spatial position can fail to provide sufficient discriminative cues.
We conclude three representative types of occlusion-induced identity (ID) consistency problems, as illustrated in Fig. \ref{fig:occlusion-induced problems}:  (a) Identity of the front object switched to the occluded object's identity; (b) Reinitialization of the occluded object after reappearance; (c) Identity swap of two objects after occlusion and trajectories intersection.

To improve the tracking robustness against occlusions and non-linear motions, recent work has tried to introduce additional motion cues in similarity evaluation \cite{cao2023observation}. Meanwhile, depth information has been proven to be effective in target set decomposition under dense occlusions in MOT \cite{liu2023sparsetrack}. 
However, to the best of our knowledge, no existing methods have tried to incorporate depth as a state into the motion model in pure motion-based 2D MOT.

\begin{figure*}[!t]
\centering
\includegraphics[width=0.7\textwidth]{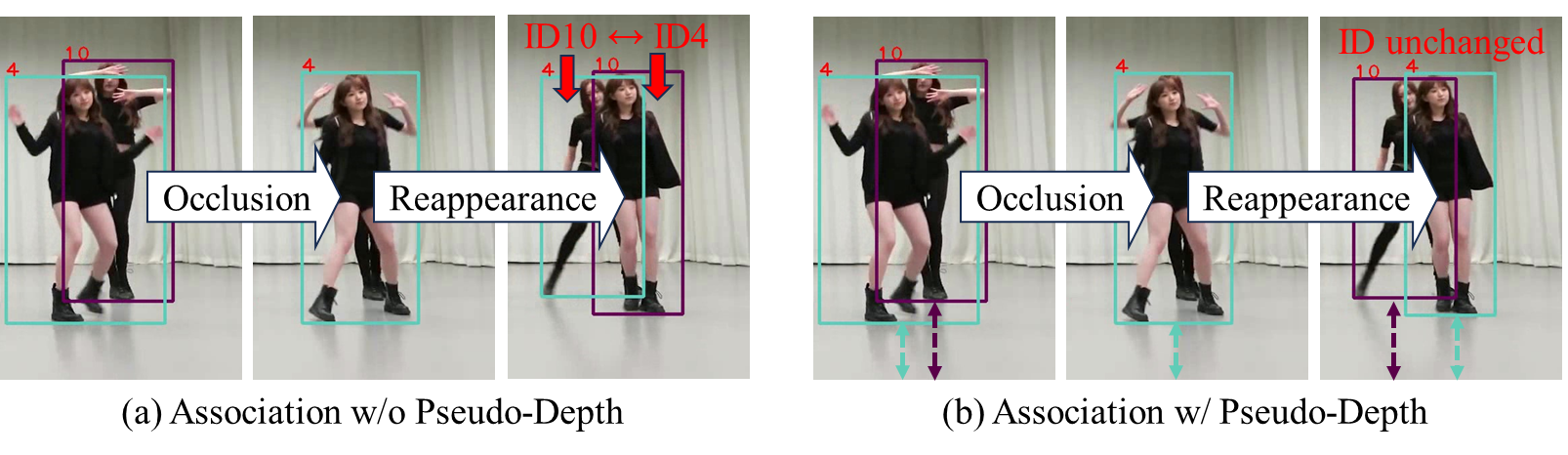}
\caption{A comparison of association without depth information and with depth information on DanceTrack \cite{sun2022dancetrack}. Bounding boxes and dashed arrows of different colors represent the location and depth of different objects. we intuitively and experimentally observe that depth information can compensate for the association failure after occlusion and reappearance.}
\label{fig:dep_illus}
\end{figure*}

In this paper, we use depth information to improve 2D MOT performance under complex scenes with dense occlusions by introducing pseudo-depth into the MOT motion model.
First, we develop a simple method to extract pseudo-depth from 2D images. With the concept of a complementary view, our pseudo-depth is robust to boundary cases. 
Next, we employ the Kalman filter (KF) \cite{kalman1960new} to model the object's motion, as it is a typical approach for motion prediction in TBD methods. Specifically, we extend the widely used KF motion state from SORT \cite{bewley2016simple} with pseudo-depth and its velocity. To achieve more accurate target localization, we design a depth-volume intersection over union (DVIoU) that uses pseudo-depth to expand the standard 2D intersection over union (IoU) \cite{rezatofighi2019generalized} similarity to 3D.
In addition, we also introduce the camera motion compensation (CMC) \cite{aharon2022bot} technique to improve the tracking quality in dynamic camera environments.
As shown in Fig. \ref{fig:dep_illus}, we experimentally find that depth information is consistent under occlusion, and can compensate for the association of 2D information.

For the implementation, we adopt OC-SORT \cite{cao2023observation} as our base method for its concise structure and strong performance. We inherent the observation-centric idea of OC-SORT and implement our designs using historical observations. Firstly, pseudo-depth computation and camera motion compensation are performed at the beginning of each frame. Secondly, our DVIoU replaces the IoU similarities in both the regular association and the recovery of lost tracklets using their historical observations (Observation-Centric Recovery, or OCR in OC-SORT). Finally, the QPDM cost is added to the cost matrix along with the DVIoU cost and the velocity consistency cost (Observation-Centric Momentum, or OCM in OC-SORT). As our focus is to introduce pseudo-depth into the MOT motion model, we name our method Pseudo-Depth SORT (PD-SORT). By integrating the above designs, PD-SORT consistently outperforms its baseline in MOT17, MOT20, and DanceTrack in most MOT metrics (see Tables \ref{tab:benchmark_dance}, \ref{tab:benchmark_mot17}, and \ref{tab:benchmark_mot20}) while remaining a simple, online, real-time, and pure motion-based tracker.

The main contributions of our work are three-fold:
\begin{itemize}
    \item We incorporate the pseudo-depth information into 2D MOT and demonstrate its effectiveness in alleviating association failures caused by occlusions and non-linear motions.
    \item We design Depth Volume IoU (DVIoU) and Quantized Pseudo-Depth Measurement (QPDM) to leverage the depth information in association, which effectively reduces the cases of association errors.
    \item We propose PD-SORT by integrating our designs into OC-SORT. PD-SORT consistently outperforms its baseline on MOT17, MOT20, and DanceTrack. This proves the generalization ability of PD-SORT across diverse MOT scenes.
\end{itemize}

The remainder of this paper is organized as follows: Section \ref{sec: rw} reviews related works on data association and the use of depth information in multi-object tracking. Section \ref{sec: method} presents our proposed tracking method. Section \ref{sec: Experiments} reports the experimental setup and evaluation results, including ablation studies and benchmark comparisons. Finally, Section \ref{sec: conclusion} concludes this paper with a summary of key contributions and potential future directions.

\section{Related Work}\label{sec: rw}
Multi-object tracking (MOT) is an essential task in the vision field that has become a hot research topic. The present MOT methods can be categorized into two types, namely the end-to-end tracking methods \cite{sun2020transtrack,meinhardt2022trackformer,zeng2022motr} and the tracking-by-detection (TBD) methods \cite{bewley2016simple,wojke2017simple,zhang2022bytetrack,cao2023observation}.
Due to its simplicity and strong performance, tracking-by-detection is the mainstream paradigm among the MOT methods.
In particular, the prevalent TBD paradigm divides MOT into two steps: detection and association. Due to the rapid development of modern deep detectors \cite{ren2015faster,redmon2018yolov3,ge2021yolox}, research in the field of MOT focuses on how to achieve more reliable association. At the same time, depth information provides key information in 3D MOT and shows its potential to improve tracking quality in 2D MOT.

\subsection{Association in 2D MOT}
To achieve reliable association, most MOT methods that follow the TBD paradigm leverage the target’s motion consistency \cite{bewley2016simple,bochinski2017high,wojke2017simple,he2021learnable,zhang2022bytetrack,cao2023observation}. The pioneering work SORT \cite{bewley2016simple} employs the Kalman filter (KF) \cite{kalman1960new} to model the target motion: at the beginning of each frame, the motion states of the targets are predicted by the KF using the linear motion assumption. Then, the IoU similarities between the predictions and the detections are calculated and used in the cost matrix for matching by the Hungarian algorithm \cite{kuhn1955hungarian}. After being successfully matched, the corresponding new detections are used to update the tracklets' KF parameters. This association pipeline of SORT is followed and improved by later TBD methods \cite{wojke2017simple,zhang2022bytetrack,cao2023observation}. To alleviate the high ID switch of SORT under occlusion, DeepSORT \cite{wojke2017simple} introduces ReID-based appearance similarity in the cost matrix. Also, it proposes an association strategy that prioritizes the tracklets with more recent successful associations. 
To effectively integrate appearance cues, SAT \cite{suljagic2022SAT} explores a deep Siamese network to extract instance-level appearance features. The obtained features are then used for similarity computation in the association stage.
Besides, appearance features extracted by deep appearance models \cite{202luo0botreid,he2023fastreid} provide effective discriminating cues that benefit tracking quality, which are exploited by later works \cite{wang2020jde,zhang2021fairmot,du2023strongsort,zhang2023stat}. To realize more reliable association, BoT-SORT \cite{aharon2022bot} modifies the KF model and uses the camera motion compensation technique to generate more accurate KF predictions while combining motion and appearance cues. Due to factors like occlusion and motion blur, low-confidence detections can also indicate the existence of targets. However, both SORT and DeepSORT perform associations for high-confidence detection results only. Therefore, ByteTrack \cite{zhang2022bytetrack} proposes a new matching cascade strategy: Once high-confidence detections have been matched, low-confidence detections and tracklets not matched with high-confidence detections are also matched. By considering all the detections, ByteTrack effectively improves the association performance of the SORT-like method. But it still has limitations when dealing with nonlinear motions and occlusions. When interruptions happen, the parameters of the Kalman filter cannot be updated due to the absence of new observations. And the KF prediction error will accumulate over time. On the other hand, the error of the observations (detections) depends on the detector, which is stable and smaller than the KF errors. Therefore, OC-SORT \cite{cao2023observation} uses the tracklets’ historical observations to compute the velocity-direction consistency with the new detections as well as to recover the interrupted tracklets. Also, after the target is reappeared, the observations before and after interruption are used to interpolate a virtual trajectory, which is then used to update the KF. Generally, the main challenge of TBD methods is the association under complex scenes, including dense objects, heavy occlusions, and nonlinear motions.
\subsection{Depth Information in MOT}
In instance-level object identification tasks, effectively leveraging scene context information can enhance the model's ability to distinguish targets \cite{bayraktar2022fast}. For the MOT task, exploring richer scene context can contribute to more robust object association. As an effective form of spatial context, depth information can refine the motion modeling of targets, thereby improving the tracker’s localization and discrimination capabilities. In 3D MOT,
AB3DMOT \cite{weng2020ab3dmot} obtains detections with depth information from a LiDAR point cloud and extends the KF to be 3D. CenterPoint \cite{yin2021center} detects object centers using a keypoint detector and estimates attributes like 3D size, orientation, and velocity. It refines these estimates using point features and simplifies the tracking to greedy closest-point matching. To obtain a comprehensive understanding of the scene, EagerMOT \cite{kim2021eagermot} fuses object observations from both 3D and 2D object detectors. 
However, in mobile device applications (e.g., smartphones), deploying depth sensors brings additional costs. Meanwhile, actual depth data obtained from depth sensors is often limited by their perception range, resulting in reduced tracking performance for distant targets.
In fact, as a projection of the 3D scene, a 2D image also implies certain depth information. 
In 2D MOT, previous work have attempted to enhance tracking performance by incorporating pseudo-depth extracted from the image signal \cite{dendorfer2022quo,quach2024depth,liu2023sparsetrack}. QuoVadis \cite{dendorfer2022quo} combines the 2D detector with a monocular depth estimator and a segmentation network to achieve trajectory forecasting from a Bird’s-Eye View (BEV). However, this method has a high model complexity.
On the other hand, DP-MOT \cite{quach2024depth} uses a geometry-based approach to estimate the depth for detected objects. Then, tracking is performed by joint use of the depth-aware motion cue and the appearance cue. Similarly, SparseTrack \cite{liu2023sparsetrack} proposes a projection rule-based method for obtaining the relative depth of targets from 2D images, which does not require training any additional networks.
Based on this pseudo-depth, the tracklets and detections are divided into subsets. Eventually, cascaded matching is performed on tracklets and detections that are at the same depth level. 
However, the aforementioned 2D MOT methods treat pseudo-depth as an auxiliary cue for constructing BEV, complementing it with appearance features, or partitioning object subsets. In contrast, we propose to integrate pseudo-depth directly into the target’s motion model as a reliable motion state, aiming at enhancing the tracker's robustness in complex scenarios with dense occlusions.

\section{Methodology}\label{sec: method}
\begin{figure*}
    \centering
    \includegraphics[width=0.82\textwidth]{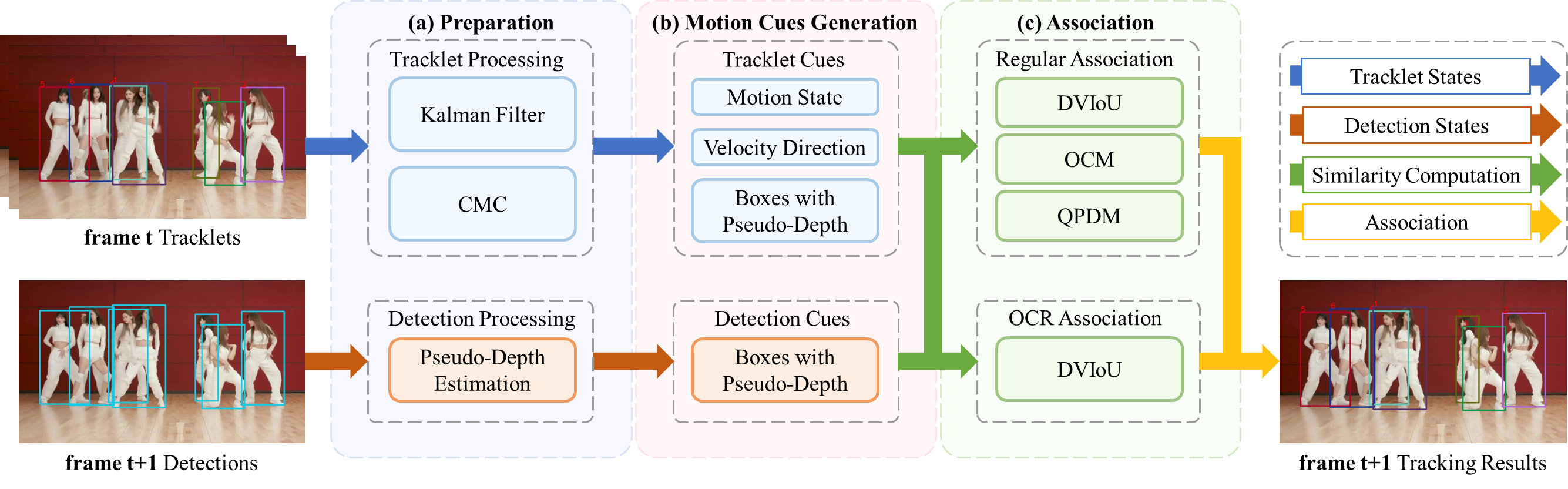}
    \caption{
    Pipeline of PD-SORT. The preparation stage estimates pseudo-depth for new detections and uses CMC to correct both motion states from KF and historical observations. For the motion cues generation, pseudo-depth is incorporated into motion states and bounding box locations for both tracklets and detections. The association stage utilizes the motion cues to compute pseudo-depth guided matching similarities in terms of DVIoU and QPDM, and the velocity consistency described by OCM to perform a two-stage association to match between tracklets and detections.
    }
    \label{fig:asso_pip}
\end{figure*}

In this section, we introduce the main components of the proposed PD-SORT, including the pseudo-depth modeling approach, the strategies to exploit depth information in the association stage, namely Depth Volume IoU (DvIoU) and Quantized Pseudo-Depth Measurement (QPDM). And the camera motion compensation (CMC) is also integrated to alleviate the camera movement problems common in MOT scenes. 
The overall pipeline is shown in Fig. \ref{fig:asso_pip}. PD-SORT produces tracking results for frame t+1 by matching detections of frame t+1 with tracklets from frame t, which comprises three core steps: (a) Preparation: CMC corrects the targets’ KF states and historical observations, and the pseudo-depth values of the detections are estimated. (b) Motion Cues Generation: The motion states in the new frame are predicted using the corrected KF states, the velocity directions are computed using historical observations, and the locations (bounding boxes and pseudo-depth) of detections and tracklets are both recorded. (c) Association: A two-stage association is performed using detection locations and tracklet cues. The first stage of regular association considers three similarities: DVIoU that computes location similarity based on KF-predicted motion states; OCM that computes velocity direction consistency with the detections; and QPDM that checks the pseudo-depth consistency. For unmatched detections and tracks, the OCR association is then performed, using the DVIoU between the detections and the tracklets' last historical observation as the association criterion to recover unmatched tracklets.
Notably, PD-SORT is developed upon OC-SORT, which retains observation-centric modules in OC-SORT (i.e., OCM, OCR, and ORU) and uses historical observations to calculate similarity.
\subsection{Pseudo-Depth Modeling}
In 2D MOT, the robustness of association relies on the estimation of the object's position, which is highly susceptible to nonlinear motions and occlusions. On the other hand, by expanding the spatial information of the object, 3D tracking that includes depth information can effectively improve the accuracy of object localization and robustness to occlusions. Meanwhile, the effectiveness of projection-based pseudo-depth in MOT tasks has been verified in previous work \cite{liu2023sparsetrack}. However, to the best of our knowledge, there’s no work that incorporates pseudo-depth as a state into the motion model for pure motion-based 2D MOT.
A key challenge lies in maintaining the accuracy of depth estimation when handling difficult targets like boundary objects. Reliable pseudo-depth estimation is essential, as it underpins the effectiveness of subsequent similarity computation modules. Moreover, appropriate pseudo-depth-based motion states to be integrated into the Kalman filter are required to ensure the discrimination ability of the motion predictions.

This revelation leads us to extend the MOT motion model by introducing pseudo-depth and its velocity, which in turn extends the 2D MOT to 3D for better processing. For the definition of pseudo-depth, as in SparseTrack \cite{liu2023sparsetrack}, we first used the projection of depth given by the distance from the target bounding box to the bottom of the image view. 
Such projection-based pseudo-depth estimation relies on the assumptions that the image capture device is above the ground plane and all objects in the scene are on the same plane. In practical tracking applications in terms of mobile device capturing, pedestrian monitoring, and in-car camera sensing, these assumptions are typically satisfied, enabling pseudo-depth estimation to provide effective guidance.
However, considering that the target bounding box may move to the boundary of the view during the tracking process, the pseudo-depth of the object may become a negative value or zero, which cannot correctly reflect the depth of the target for modules using depth values directly, influencing the subsequent pseudo-depth-based calculation.

\begin{figure}
    \centering
    \includegraphics[width=1.00\columnwidth]{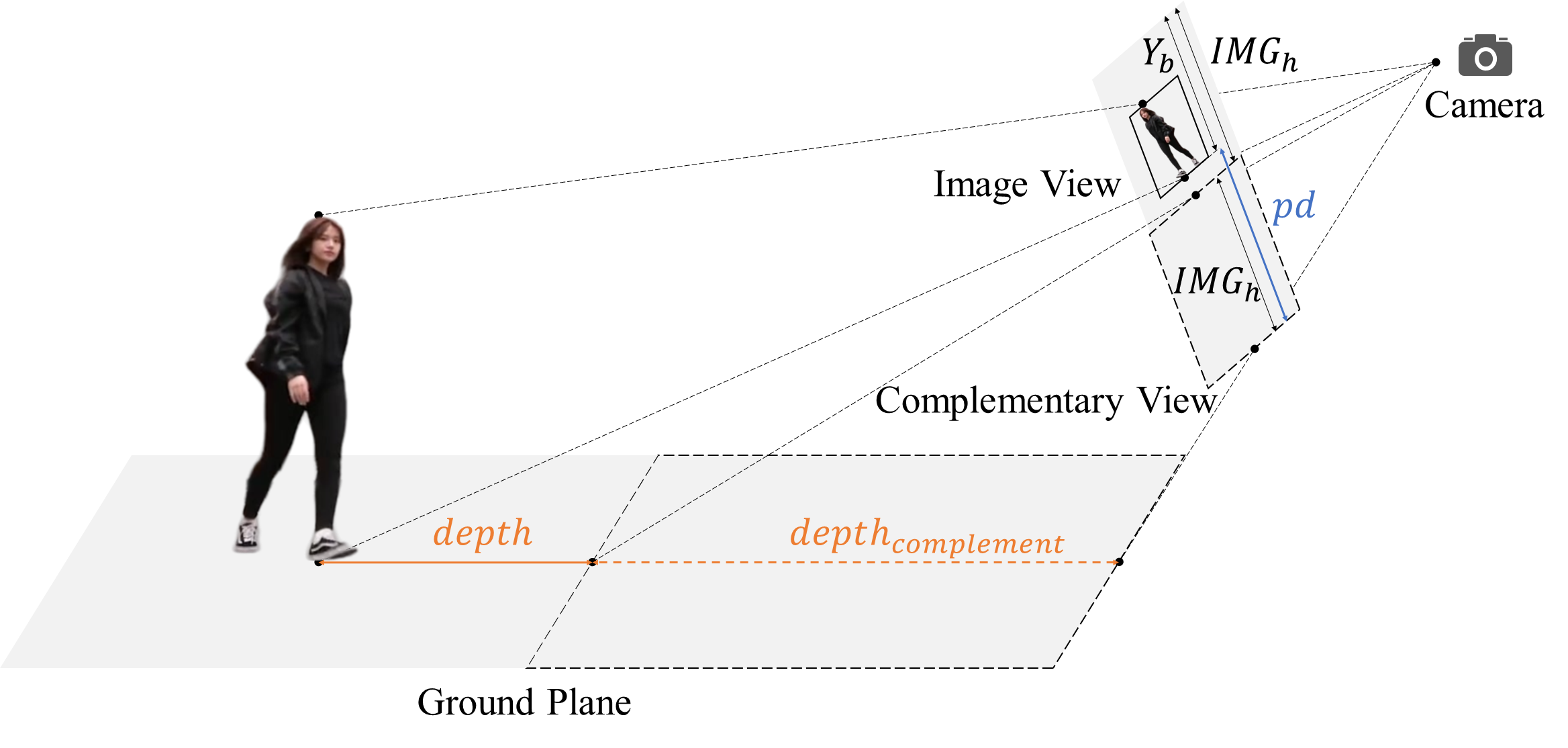}
    \caption{Illustration of our pseudo-depth. The orange double-arrow line represents the real depth on the ground plane ($depth$), 
    the dashed orange double-arrow line represents the length that corresponds to the pseudo-depth in the complementary view on the ground plane ($depth_{complement}$),
    and the blue double-arrow line represents the pseudo-depth obtained by projecting the real depth onto the view plane with both the real image view and the complementary view ($pd$).}
    \label{fig:illus-pd}
\end{figure}

Therefore, we propose a novel pseudo-depth based on the complementary view. By expanding a complementary view of the same size and below the real image view, we define the pseudo-depth as the distance from the bottom of the target bounding box to the bottom of the complementary view, and our pseudo-depth \(pd\) is computed as in Eq. \ref{eq:pdcal}.
\begin{equation}
     pd\ =\ 2\times{IMG}_{h}-Y_{b}   
    \label{eq:pdcal}
\end{equation}
Here, \({IMG}_{h}\) is the height of the real view, \(Y_{b}\) is the coordinate value of the bottom of the target bounding box along the y-axis. 
The visualization of the ground plane real depth $depth$ and our pseudo-depth $pd$ is shown in Fig. \ref{fig:illus-pd}.
For objects whose size is within the real view, such pseudo-depth using complementary view can correctly reflect the depth information.

Based on the proposed pseudo-depth, we extend the standard KF in SORT with two additional states: the target’s pseudo depth \(pd\) and its velocity component \(v_{pd}\). The standard Kalman filter states in SORT are shown in Eq. \ref{eq:oristate}. 
\begin{equation}
    X=[x_c,\ y_c,\ s,\ r,\ v_x,\ v_y,\ v_s]
    \label{eq:oristate}
\end{equation}
Here, \((x_c,\ y_c)\) is the coordinate of the target’s bounding box center, s and r are the area and aspect ratio of the target’s bounding box. $v_x,\ v_y,\ v_s$ are the velocity components for $x_c,\ y_c,\ s$ respectively. By introducing two new states, \(pd\) and \(v_{pd}\), the KF state is revised to be as in Eq. \ref{eq:newstate}.
\begin{equation}
    X=[x_c,\ y_c,\ pd,\ s,\ r,\ v_x,\ v_y,\ v_{pd},\ v_s]
    \label{eq:newstate}
\end{equation}

\subsection{Depth Volume IoU}
To utilize the depth information in location consistency evaluation, we extend the 2D IoU similarity to 3D by introducing the concept of depth volume. Given two object observations $b^1= (x_1^1,\ y_1^1,\ x_2^1,\ y_2^1,\ {pd}^1)$ and $b^2= (x_1^2,\ y_1^2,\ x_2^2,\ y_2^2,\ {pd}^2)$, where \((x_1^{1/2},\ y_1^{1/2})\), \((x_2^{1/2},\ y_2^{1/2})\), and \(pd^{1/2}\) represent the top-left corner, bottom right corner, and the pseudo-depth, respectively. We give the definition of the depth volume of the intersection between the two objects, $V^{\text{inter}}$, as in Eq. \ref{eq:vinter}.
\begin{equation}
    \left\{\begin{array}{c}
    V^{\text{inter}}=w^{\text{inter}} \cdot h^{\text{inter}} \cdot pd^{\text {inter}} \hfill\\
    w^{\text{inter}}=\min \left(x_{2}^{1},\ x_{2}^{2}\right)-\max \left(x_{1}^{1}-x_{1}^{2}\right) \hfill\\
    h^{\text{inter}}=\min \left(y_{2}^{1},\ y_{2}^{2}\right)-\max \left(y_{1}^{1}-y_{1}^{2}\right) \hfill\\
    pd^{\text{inter}}=\min \left(pd^{1},\ p d^{2}\right)\hfill\end{array}\right.
    \label{eq:vinter}
\end{equation}
Here, $w^{inter}$ and $h^{inter}$ are the width and height of the intersection box area. Meanwhile, we define the pseudo-depth of the intersection, $pd^{inter}$, as the smaller value of the pseudo-depths of the two objects. Similarly, we can obtain the depth volumes of two objects, $V^1$ and $V^2$, as in Eq. \ref{eq:vol}.

\begin{equation}
    \left\{\begin{array}{c}
    V^{1/2}=w^{1/2}\cdot h^{1/2} \cdot{pd}^{1/2} \hfill\\
    w^{1/2}=x_2^{1/2}-x_1^{1/2},\ h^{1/2}=y_2^{1/2}-y_1^{1/2}\hfill\end{array}\right.
    \label{eq:vol}
\end{equation}

Furthermore, to achieve more robust distinguishing between objects, we introduce depth volume IoU (DVIoU) by using the volume metric, as shown in Eq. \ref{eq:dviou}.
\begin{equation}
    DVIoU=\frac{V^{inter}}{V^1+V^2-V^{inter}}
    \label{eq:dviou}    
\end{equation}

\begin{figure}
    \centering
    \includegraphics[width=1.00\columnwidth]{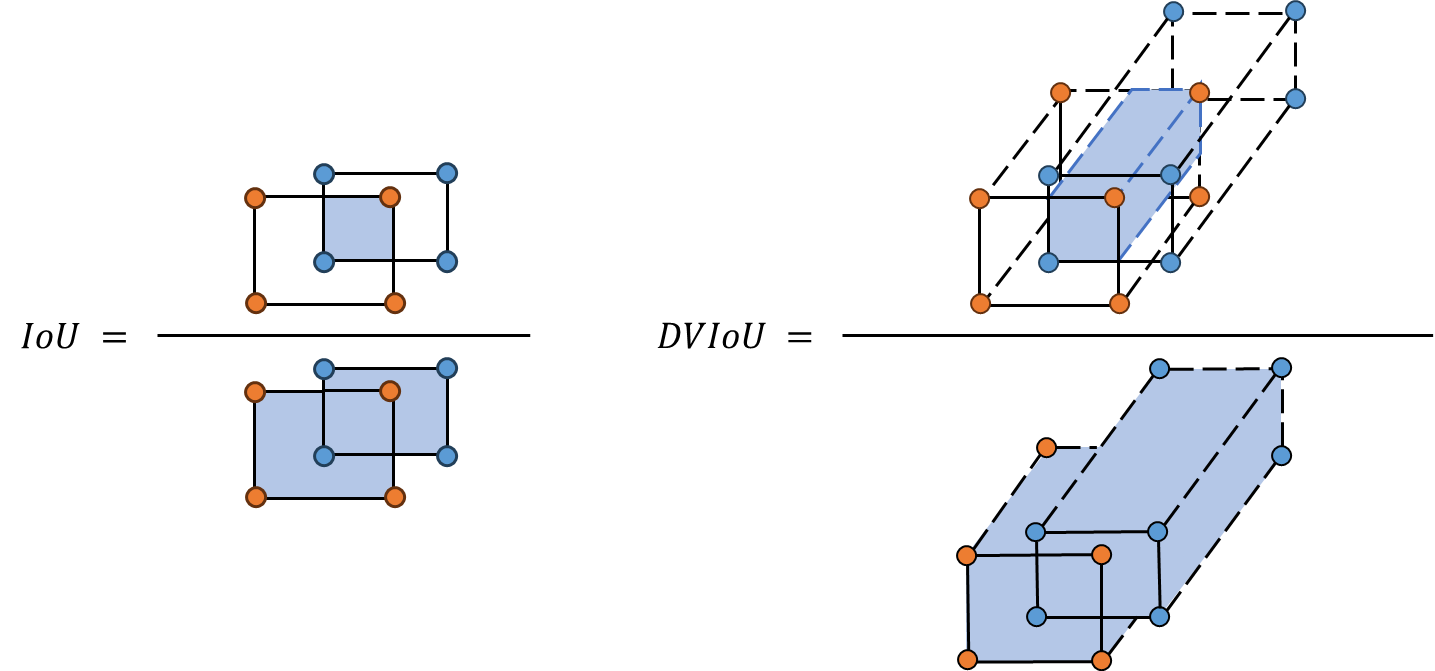}
    \caption{Illustration of IoU and DVIoU. By integrating pseudo-depth (the extra dimension represented by the dashed line in the figure), area-based standard 2D IoU is extended to volume-based DVIoU.}
    \label{fig:dviou_illus}
\end{figure}

The comparison between standard IoU and DVIoU is visually represented in Fig. \ref{fig:dviou_illus}. 
By integrating the depth to modulate the IoU similarity, not only the robustness of target location consistency measurement is improved, but also the extra discrimination information provided by the depth cue benefits the overall association accuracy.

\subsection{Quantized Pseudo-Depth Measurement }
Occlusions can harm the reliability of the pseudo-depth, which in turn leads to a decrease in tracking accuracy. On the other hand, in successive frames, the relative depth of the object with respect to other objects fluctuate only in narrow intervals. Therefore, we propose a quantized pseudo-depth cost to better utilize the pseudo-depth to guide the association. 

For each frame, we find the minimum pseudo-depth value of all detected objects in the frame \(pd_{min}\) and the maximum value \(pd_{max}\). Then, divide the interval \([pd_{min},\ pd_{max}]\) into \(interval_{num}\) sub-intervals uniformly; each sub-interval is assigned with an interval depth (in this paper, the interval depth is defined as the upper limit of the sub-interval after min-max normalization). After that, the interval depths are assigned to the objects according to the sub-intervals they are in. The interval depth for the \(i^{th}\) \((i=0,\ 1,\ …,\ {interval}_{num}-1)\) sub-interval is computed as in Eq. \ref{eq:interd}.
\begin{equation}
    \left\{\begin{array}{c}
    interdepth_i={[(i+1) \times len}_{interval}]\ / \ len_{total}\\
    {len}_{interval}=len_{total}\ / \ {interval}_{num}\hfill\\
    len_{total}=pd_{max}-pd_{min}\hfill\end{array}\right.
    \label{eq:interd}
\end{equation}

Next, the interval depth is computed for the last historical observation of each tracklet in the same manner. Finally, the quantized pseudo-depth cost \(C_{QPD}\) is computed as the absolute difference between the interval depths of the new detections, \(interdepth_{dets}\), and the interval depths of the tracklets, \(interdepth_{tracks}\), as shown in Eq. \ref{eq:costqpd}.
\begin{equation}
    C_{QPD}=abs\left(interdepth_{tracks}-interdepth_{dets}\right)
    \label{eq:costqpd}
\end{equation}

Then the pseudo-depth difference between the tracklets and new detections can be evaluated by their interval depth values. Compared to directly calculating the difference in pseudo-depth, using the proposed interval depth between detections and tracklets reduces the depth estimation error caused by partial occlusions, thus improving the robustness of pseudo-depth utilization. Meanwhile, interval depth-based cost computation helps to alleviate the association error caused by the velocity direction consistency evaluation when the object is steering, which further improves the algorithm’s performance against nonlinear motions. Finally, the pseudo-code of the QPDM algorithm is given in Algorithm \ref{alg:qpdm}.

\RestyleAlgo{ruled}
\SetKwComment{Comment}{/* }{ */}
\begin{algorithm}[t]
\caption{The pseudocode of QPDM.}\label{alg:qpdm}
\DontPrintSemicolon
\SetKwInput{KwInput}{Input}                
\SetKwInput{KwOutput}{Output}              
\KwInput{number of sub-intervals $interval_{num}$, pseudo-depth set of tracklets’ previous observations $pd_{obs}$, pseudo-depth set of new detections $pd_{dets}$}
\KwOutput{The pseudo-depth cost matrix between tracklets and detections $C_{QPD}$}
$len_{obs} \gets max(pd_{obs}) - min(pd_{obs})$\;
$pd_{obs} \gets (pd_{obs}-min(pd_{obs})/len_{obs}$\;
$min_{previous} \gets 1$\;
\Comment{Compute interval depth for previous observations}
\For{$inter \leftarrow 0$ \KwTo $interval_{num} - 1$}{
    $min_{current} \gets 1-(inter+1)/interval_{num}$\;
    $inter_{obs}^{depth}[min_{previous}\leq pd_{obs} \leq min_{current}] \gets min_{current} + 1/interval_{num}$\;
    $min_{previous} \gets 1$\;
}
$len_{dets} \gets max(pd_{dets}) - min(pd_{dets})$\;
$pd_{dets} \gets (pd_{dets}-min(pd_{dets})/len_{dets}$\;
$min_{previous} \gets 1$\;
\Comment{Compute interval depth for new detections}
\For{$inter \leftarrow 0$ \KwTo $interval_{num} - 1$}{
    $min_{current} \gets 1-(inter+1)/interval_{num}$\;
    $inter_{dets}^{depth}[min_{previous}\leq pd_{dets} \leq min_{current}] \gets min_{current} + 1/interval_{num}$\;
    $min_{previous} \gets 1$\;
}
$C_{QPD} \gets abs(inter_{obs}^{depth} - inter_{dets}^{depth})$\;
\KwRet $C_{QPD}$\;
\end{algorithm}

\subsection{Camera Motion Compensation}
In our association method, the motion information is consisting of 3 parts: the DVIoU similarity, the OCM velocity-direction consistency, and the quantized pseudo-depth loss. Among them, both DVIoU and OCM are sensitive to the position information of the target. For example, for the DVIoU, the depth volume is the product of pseudo-depth and 2D box area. Here, the pseudo-depth is a relative position information robust to camera motion, but the 2D bounding box overlap is sensitive to position drift. Once the position of either the previous observation or the current detection drifts, the overlap area will change largely and can lead to incorrect association. Meanwhile, OCM relies on the center point coordinates of historical observations to calculate the velocity direction, which is also sensitive to the offset of the target center point. Thus, the accuracy of the target position is essential for association quality.

However, when the camera moves, the position of the target in the view will also shift, which affects the association result. To this end, we introduce CMC before KF’s prediction step for more robust tracklet-detection association in the coming frame. Specifically, we use the OpenCV \cite{bradski2000opencv} implementation of the Video Stabilization module with affine transformation to generate transforms using key point extraction \cite{shi1994good}, sparse optical flow \cite{bouguet2001pyramidal}, and RANSAC \cite{fischler1981random}, as in previous work \cite{aharon2022bot}. Given a scale and rotation matrix \(M\in R^{2\times2}\) and a translation \(T\in R^{2\times1}\), we correct the camera motion of the KF state and the target historical observation as follows.

\subsubsection{KF State Correction}
The KF state \(X\) of our method is depicted in Eq. \ref{eq:newstate}, where $(x_c,\ y_c)$ is the center coordinate of the target, \(pd\) is the pseudo-depth of the target, \(s,\ r\) are the bounding box area and aspect ratio, respectively. And \(v_x,\ v_y,\ v_{pd},\ v_s\) are the corresponding velocities. We apply the CMC to the state \(X\) and the KF’s covariance matrix \(P\) following Eq. \ref{eq:cmc1}.
\begin{equation}
    \left\{\begin{array}{c}X\left[0:2\right]=M  X\left[0:2\right]+T \hfill\\
    X\left[5:7\right]=M  X\left[5:7\right]+T \hfill\\
    P\left[0:2,\ 0:2\right]=M  P\left[0:2,\ 0:2\right]  M^T \hfill\\
    P\left[5:7,\ 5:7\right]=M  P\left[5:7,\ 5:7\right]  M^T \hfill\end{array}\right.
    \label{eq:cmc1}
\end{equation}
\subsubsection{Historical Observation Correction}
The three modules in OC-SORT, OCM, ORU and OCR, use the center positions of historical observations to compute the direction of target motion, generate virtual positions when trajectory interruptions and reappearances happen, and match with KF predictions, respectively. Thus, we also apply CMC to the tracklets’ historical observations. Supposing the center position of a historical observation is \(p_c=(x_c,{\ y}_c)\), the CMC is performed as Eq. \ref{eq:cmc2}.
\begin{equation}
    p_c=M {p}_c + T
    \label{eq:cmc2}
\end{equation}

By correcting the target center position in Kalman filter state vectors and historical observations, we reduce the error in the DVIoU computation, while making the velocity-direction consistency computation of the OCM module more accurate, thus improving the overall association accuracy.
\subsection{Algorithm Overall Framework}
For new detections in each frame, OC-SORT performs a two-stage association: the first stage of regular association using the IoU and the velocity consistency (OCM), followed by a second stage to recover the lost tracklets using the IoU only (OCR).
PD-SORT follows the association flow of OC-SORT and additionally adds pseudo-depth cues to the associations. First, the QPDM module, which directly leverages pseudo-depth, is introduced into the regular association. Meanwhile, the conventional IoU similarities used in both rounds of associations are replaced with the proposed DVIoU, which also uses pseudo-depth. Eventually, the composition of the final cost matrix is shown in Eq. \ref{eq:costall}.
\begin{equation}
    C\ =\ C_{DVIoU}+\lambda_1C_{QPD}+\ \lambda_2C_{OCM}
    \label{eq:costall}
\end{equation}
Here, $C_{DVIoU}$ is the opposite of the DVIoU between KF predictions and the detections. $C_{QPD}$ is the QPDM cost. $C_{OCM}$ is inherent from OC-SORT, which is the velocity direction consistency difference between historical observations and new detections. $\lambda_1$ and $\lambda_2$ are two weighting factors.
The detailed pseudo-code for PD-SORT is shown in Algorithm \ref{alg:pdsort}.

\RestyleAlgo{ruled}
\SetKwComment{Comment}{/* }{ */}
\begin{algorithm}[t]
\caption{The pseudocode of PD-SORT.}\label{alg:pdsort}
\DontPrintSemicolon
\SetKwInput{KwInput}{Input}                
\SetKwInput{KwOutput}{Output}              
\KwInput{Detections $Z=\{z_k^i\mid1\le k\le T,1\le i\le N_k\}$; Kalman Filter $KF$; threshold to remove untracked tracks $t_{expire}$}
\KwOutput{The set of tracklets $\mathcal{T} ={\tau_i}$}
Initialization: $\mathcal{T} \gets \emptyset$\;

\For{$timestep t \leftarrow 1$ \KwTo $T$}{
    \Comment{Step 1: regular association to match detections with tracklets}
    $Z_t \gets \{z_t^1,\ ...,\ z_t^{N_t}\}^T$\;
    Apply CMC to last observations and last KF states for all tracklets in $\mathcal{T}$\;
    ${\hat{X}}_t \gets \{{\hat{x}}_t^1,\ ...,\ {\hat{x}}_t^{\lvert \mathcal{T} \rvert}\}$ \tcc*{Estimations by KF.predict}
    $Z\gets$ Historical observations on the existing tracks\;
    $C_t \gets C_{DVIoU}({\hat{X}}_t,\ Z_t)+\lambda_1C_{QPD}(Z,\ Z_t)+\lambda_2C_{OCM}(Z,\ Z_t)$\;
    Linear assignment by Hungarians with cost $C_t$\;
    $\mathcal{T}_t^{matched} \gets$ tracklets matched to a detection\;
    $\mathcal{T}_t^{remain} \gets$ tracklets not matched to a detection\;
    $Z_t^{remain} \gets$ detections not matched to any tracklet\; 
    \Comment{Step 2: perform OCR to find lost tracklets back}
    $Z^{\mathcal{T}_t^{remain}} \gets$ last matched detection of tracklets in $\mathcal{T}_t^{remain}$\;
    $C_t^{remain} \gets C_{DVIoU}(Z^{\mathcal{T}_t^{remain}},\ Z_t^{remain})$\;
    Linear assignment by Hungarians with cost $C_t^{remain}$\;
    $Z_t^{unmatched} \gets$ detection unmatched to tracklets\;
    update $\mathcal{T}_t^{matched}$ and $\mathcal{T}_t^{remain}$\;
    \Comment{Step 3: update states of matched tracklets}
    \For{$\tau$ in $\mathcal{T}_t^{matched}$}{
    perform ORU in OC-SORT to update KF.parameters\;
    }
    \Comment{Step 4: initialize and remove tracklets}
    $\mathcal{T}_t^{new} \gets$ new tracklets generated from $Z_t^{unmatched}$\;
    \For{$\tau$ in $\mathcal{T}_t^{remain}$}{
        $\tau.untracked \gets \tau.untracked + 1$\;
    }
    $\mathcal{T}_t^{reserved} \gets$ \{$\tau \mid \tau \in \mathcal{T}_t^{remain}$ and $\tau.untracked < t_{expire}$\}\;
    $\mathcal{T} \gets \{\mathcal{T}_t^{new},\ \mathcal{T}_t^{matched},\ \mathcal{T}_t^{reserved}\}$    
}
\KwRet $\mathcal{T}$\;
\end{algorithm}

\section{Experiments}
\label{sec: Experiments}
\subsection{Datasets and Metrics}
\subsubsection{Datasets}
We evaluated our model under the “private detection” protocol on multiple MOT datasets, including DanceTrack \cite{sun2022dancetrack}, MOT17 \cite{milan2016mot16} and MOT20 \cite{dendorfer2020mot20}.
The MOT17 dataset contains 7 training videos and 7 test videos, in which the targets have different appearances and nearly linear motions.
The MOT20 dataset contains 4 training videos and 4 test videos, where the scenes are similar to those in MOT17 but are more crowded.
DanceTrack is a recently proposed dataset where targets have similar appearances, nonlinear motions, and frequent occlusions. DanceTrack consists of 40 training videos, 25 validation videos, and 35 test videos, with more frames to comprehensively reflect the tracker's performance. Meanwhile, the detection task in DanceTrack is relatively simple, making it ideal for association quality evaluation.
Considering the characteristics of the above datasets and the goal of improving association ability in scenes with occlusions and nonlinear motions, we prioritize the comparison results on the DanceTrack dataset. Meanwhile, the generalization ability of our tracker is evaluated on both MOT17 and MOT20.
\subsubsection{Metrics}
We take HOTA \cite{luiten2021hota} as our main metric as it provides a comprehensive evaluation of tracking quality in terms of both the detection accuracy and the association accuracy. Besides, we also adopt MOTA, AssA, IDF1, and other commonly used metrics to reflect the performance of tracking algorithms from different aspects \cite{luiten2021hota,bernardin2008mota,ristani2016idf1}. Here, MOTA combines false positives, missed targets, and identity switches (IDs), and focuses on the detection performance, while AssA and IDF1 reflect the ability of associations.
\subsubsection{Implementation Details}
To maintain a fair comparison, we use the same detector as previous works. Specifically, our detection model is YOLOX \cite{ge2021yolox} with publicly available weights from our baseline OC-SORT. The weight factor for the QPDM cost is 0.2 in both DanceTrack and MOT17, and 0.36 in MOT20, where our QPDM is more beneficial. For simplicity, we divide the pseudo depth into 8 subintervals in QPDM for all three benchmarks. The OCM cost weights are 0.2 in DanceTrack and MOT17, and 0.04 in MOT20. The IoU thresholds during association are 0.3 for DanceTrack and MOT17, and 0.35 for MOT20.  Following the common practice of SORT-like methods, we set the detection confidence threshold at 0.4 for MOT20 and 0.6 for other datasets. All experiments are performed on an Intel i5-13600K CPU @ 2.60 GHz and a single NVIDIA GeForce RTX 4090 GPU.
\subsection{Benchmarks Evaluation}
We compare our PD-SORT with state-of-the-art trackers on the test sets of DanceTrack, MOT17, and MOT20, as shown in Tables \ref{tab:benchmark_dance}, \ref{tab:benchmark_mot17}, and \ref{tab:benchmark_mot20}, respectively. Note that all of the test results are evaluated on official websites. 
\subsubsection{Baseline Selection}
OC-SORT is a motion-based, SORT-like tracker. As shown in Table 1, OC-SORT shows leading tracking performance on the DanceTrack dataset in terms of HOTA, IDF1, AssA, and AssR compared to previous methods. For methods with comparable performance, StrongSORT++ and STAT integrate additional appearance feature components, and SparseTrack employs a subset decomposition and cascading strategy. These models involve more sophisticated designs and high computational costs. In contrast, OC-SORT achieves competitive performance while maintaining a simple, extensible architecture and real-time tracking speed. Therefore, we select OC-SORT as our baseline method.
\subsubsection{DanceTrack}
We report experimental results on the DanceTrack in Table \ref{tab:benchmark_dance} to evaluate PD-SORT under complex scenes with similar appearances, nonlinear motions, and frequent occlusions. Compared with its baseline OC-SORT, PD-SORT has made considerable progress in most core metrics (i.e., +3.6 HOTA, +0.2 DetA, +1.9 AssA, +2.9 IDF1). Specifically, it achieves a significantly higher HOTA than previous trackers and exceeds the base method by 6.6\%, which shows the strength of depth cues in improving the overall tracking quality. Also, the improvements on both AssA (+1.9) and IDF1 (+2.9) metrics are substantial, which further indicates the benefit of depth information to the association.

The underlying reason is that previous methods leverage pure 2D motion information, making it difficult to distinguish objects with highly overlapped bounding boxes, which often happens in occlusion cases. Nevertheless, we use pseudo-depth to provide additional cues for association. By integrating our proposed pseudo-depth modules, the occlusion-induced problems are effectively alleviated, demonstrating the robustness of PD-SORT in handling challenging scenes with diverse motions and occlusions, as in DanceTrack.
For the computational efficiency, we test the frames per second (FPS) of our method (28.7 FPS) and the baseline (35.1 FPS) on on the same device. With only 6.4 FPS lower, the tracking performance improved significantly.
    \begin{table*}[t]
        \centering
        \caption{Results on DanceTrack test set. SORT, DeepSORT, ByteTrack, StrongSORT++, SparseTrack, STAT, OC-SORT and our method share the same detections. }
        \label{tab:benchmark_dance}
        \resizebox{0.66\textwidth}{!}{
        \begin{tabular}{l | c | c c c c c}
        \hline
        \hline
             \textbf{Tracker} & \textbf{Reference} & \textbf{HOTA↑} &\textbf{DetA↑} & \textbf{AssA↑} & \textbf{MOTA↑} & \textbf{IDF1↑} \\
             \hline
             TraDeS \cite{wu2021track}&CVPR21 &43.3 & 74.5 & 25.4 & 86.2 & 41.2\\
             MOTR \cite{zeng2022motr}&ECCV22 & 54.2 & 73.5 & 40.2 & 79.7 & 51.5\\
             GTR \cite{zhou2022global}&CVPR22 & 48.0 & 72.5 & 31.9 & 84.7 & 50.3\\
             CenterTrack \cite{zhou2020tracking}&ECCV20 & 41.8 &	78.1 &	22.6 &	86.8 &	35.7\\
             FairMOT \cite{zhang2021fairmot}&IJCV21 & 39.7 & 66.7 & 23.8 & 82.2 & 40.8\\
             QDTrack \cite{pang2021quasi}&CVPR21 & 45.7 & 72.1 & 29.2 & 83.0 & 44.8\\
             TransTrack \cite{sun2020transtrack}&arXiv20 & 45.5 & 75.9 & 27.5 & 88.4 & 45.2\\
             SORT \cite{bewley2016simple}&ICIP16 & 47.9 & 72.0 & 31.2 & 91.8 & 50.8\\
             DeepSORT \cite{wojke2017simple}&ICIP17 & 45.6 & 71.0 & 29.7 & 87.8 & 47.9\\
             ByteTrack \cite{zhang2022bytetrack}&ECCV22 & 47.3 & 71.6 & 31.4 & 89.5 & 52.5\\
             StrongSORT++ \cite{du2023strongsort}&TMM23 & 55.6 & 80.7 & 38.6 & 91.1 & 55.2\\
             SparseTrack \cite{liu2023sparsetrack}&arXiv23 & 55.5 & 78.9 & 39.1 & 91.3 & 58.3\\
             STAT \cite{zhang2023stat}&TMM23 & 57.4 & 80.8 & 40.9 & 91.5 & 59.2\\
             OC-SORT \cite{cao2023observation}&CVPR23 & 54.6 & 80.4 & 40.2 & 89.6 & 54.6\\
             \textbf{PD-SORT}& Ours& 58.2 & 80.6 & 42.1 & 89.6 & 57.5\\
        \hline
        \hline
        \end{tabular}
        }
    \end{table*}
\subsubsection{MOT17 \& MOT20}
In addition to DanceTrack, we also evaluate our method on the general MOT Challenge datasets under private detection mode. For the results in MOT17 and MOT20, we inherit the linear interpolation from baseline methods for a fair comparison.
The results of the MOT17 test set are presented in Table \ref{tab:benchmark_mot17}. Compared with OC-SORT, PD-SORT made considerable progress in most core metrics (i.e., +0.8 HOTA, +1.3 MOTA, +1.7 IDF1, +0.9 AssA). The results show that PD-SORT can still achieve performance improvements on linear motion scenes. Generally, the results on MOT17 indicate that PD-SORT can generalize well in scenes with simple motions.

We also report the performance of PD-SORT on MOT20 in Table \ref{tab:benchmark_mot20}. Compared with OC-SORT, PD-SORT achieves performance gains in several core metrics (i.e., +0.5 HOTA, +0.8 IDF1, +1.1 AssA). MOT20 has more crowded scenes and a longer video length than MOT17. Such characteristics pose the challenges of long-term tracking and more severe occlusions for MOT. The results on MOT20 further demonstrate the good generalization ability of PD-SORT and its robustness against dense scenes with occlusions.
    \begin{table*}[t]
        \centering
        \caption{Results on MOT17 test set with the private detections. ByteTrack, STAT, OC-SORT and our method share the same detections.}
        \label{tab:benchmark_mot17}
         \resizebox{0.66\textwidth}{!}{
         \begin{tabular}{l | c | c c c c c c}
        \hline
        \hline
             \textbf{Tracker} & \textbf{Reference} & \textbf{HOTA↑} & \textbf{MOTA↑} & \textbf{IDF1↑} &  \textbf{IDs↓} & \textbf{AssA↑} & \textbf{AssR↑}\\
             \hline
             TrackFormer \cite{meinhardt2022trackformer}& CVPR22 & 57.3 & 74.1 & 68.0 & 2829 & - & -\\
             MOTR \cite{zeng2022motr}& ECCV22 & 57.8 & 73.4 & 68.6 & 2439 & 55.7 & -\\
             MeMOT \cite{cai2022memot}& CVPR22 & 56.9 & 72.5 & 69.0 & 2724 & 55.2 & -\\
             MOTFR \cite{2022MOTFR}& TCSVT22 & 61.8 & 74.4 & 76.3 & 2652 & 62.6 & 67.8\\
             MAA \cite{2022maa}& WACVW22 & 62.0 & 79.4 & 75.9 & 1452 & 60.2 & 67.3\\
             MOTRv2 \cite{zhang2023motrv2}& CVPR23 & 62.0 & 78.6 & 75.0 & - & 60.6 & -\\
             MO3TR-PIQ\cite{2023mo3tr}& TPAMI23 &57.3 & 72.3 & 69.0 & 2200 & - & -\\
             FairMOT \cite{zhang2021fairmot}& IJCV21 & 59.3 & 73.7 & 72.3 & 3303 & 58.0 & 63.6\\
             QDTrack \cite{pang2021quasi}& CVPR21 & 53.9 & 68.7 & 66.3 & 3378 & 52.7 & 57.2\\
             CorrTracker \cite{wang2021multiple}& CVPR21 & 60.7 & 76.5 & 73.6 & 3369 & 58.9 & 64.4\\
             ByteTrack \cite{zhang2022bytetrack}& ECCV22 & 63.1 & 80.3 & 77.3 & 2196 & 62.0 & 68.2\\
             STAT \cite{zhang2023stat}& TMM23 & 63.7 & 78.7 & 79.0 & 2754 & 63.4 & 70.6\\
             OC-SORT \cite{cao2023observation}& CVPR23 & 63.2 & 78.0 & 77.5 & 1950 & 63.2 & 67.5\\
             \textbf{PD-SORT}& Ours & 63.9 & 79.3 & 79.2 & 1062 & 64.1 & 69.4\\
        \hline
        \hline
        \end{tabular}
         }
    \end{table*}
    \begin{table*}[t]
        \centering
        \caption{Results on MOT20 test set with the private detections. ByteTrack, GHOST, STAT, OC-SORT and Ours share the same detections.}
        \label{tab:benchmark_mot20}
         \resizebox{0.66\textwidth}{!}{
        \begin{tabular}{l | c | c c c c c c}
        \hline
        \hline
             \textbf{Tracker} & \textbf{Reference} & \textbf{HOTA↑} & \textbf{MOTA↑} & \textbf{IDF1↑} &  \textbf{IDs↓} & \textbf{AssA↑} & \textbf{AssR↑}\\
             \hline
             MeMOT \cite{cai2022memot}&CVPR22& 54.1 & 63.7 & 66.1 & 1938 & 55.0 & -\\
             MOTRv2 \cite{zhang2023motrv2}&CVPR23& 61.0 & 76.2 & 73.1 & - & 59.3 & - \\
             TransMOT \cite{chu2023transmot}&WACV23& 61.9 & 77.5& 75.2 & 1615 & 60.1& 66.3\\
             RelationTrack \cite{relationTrack} &TMM23& 56.5 & 67.2 & 70.5 & 4243 & 56.4 & 60.3\\
             FairMOT \cite{zhang2021fairmot}&IJCV21& 54.6 & 61.8 & 67.3 & 5243 & 54.7 & 60.7\\
             MOTFR \cite{2022MOTFR}&TCSVT22& 57.2 & 69.0 & 71.7 & 3648 & 57.1 & 62.6\\
             MAA \cite{2022maa}&WACVW22& 57.3 & 73.9 & 71.2 & 1331 & 55.1 & 61.1\\
             CSTrack \cite{2022CSTrack}&TIP22& 54.0 & 66.6 & 68.6 & 3196 & 54.0 & 57.6\\
             ByteTrack \cite{zhang2022bytetrack}&ECCV22& 61.3 & 77.8 & 75.2 & 1223 & 59.6 & 66.2\\
             GHOST \cite{2023ghost}&CVPR23& 61.2&73.7&75.2&1264&-&-\\
             STAT \cite{zhang2023stat}&TMM23& 62.5 & 75.5 & 76.4 & 975 & 62.8 & 68.2\\
             OC-SORT \cite{cao2023observation}&CVPR23& 62.1 & 75.5 & 75.9 & 913 & 62.0 & 67.5\\
             \textbf{PD-SORT}&Ours& 62.6 & 75.4 & 76.7 & 908 & 63.1 & 68.4 \\
        \hline
        \hline
        \end{tabular}
        }                
    \end{table*}

\subsection{Ablation Study}
\subsubsection{Component Ablation}
We perform ablation studies on the validation set of DanceTrack to evaluate the impact of each module in the proposed PD-SORT under complex occlusion scenes.
To achieve a valid assessment, we use the same detection model and weights as the base method, OC-SORT, across all experiments. Also, the parameter settings follow those in the baseline.
Table \ref{tab:ablation_all} presents the contribution of each module by progressively adding modules to the base method.
By correcting the position states, the CMC module benefits other modules for more accurate motion estimation in dynamic camera scenes. 
Notably, nonlinear object motions and occlusions happen frequently in DanceTrack. In such situations, the depth information becomes a reliable cue to compensate for the cases where pure 2D association fails. 
Thus, with proper strategies to leverage pseudo-depth in the association, both DVIoU and QPDM are effective in scenes like DanceTrack.
DVIoU modulates the box similarities of the objects with pseudo-depth, which is stable and rich in discriminative information while having no negative impact on the model. 
Particularly, the QPDM module directly uses the pesudo-depth to guide the association and achieves significant performance gains in DanceTrack. This also indicates that pseudo-depth quantitation is a robust technique to handle occlusions with nonlinear motions.
Additionally, scenes in DanceTrack show long durations, which are longer than conventional datasets like MOT17. The effectiveness of DVIoU and QPDM on the dataset also shows the potential of the pseudo-depth-based method for long-term MOT.
In general, the results in Table \ref{tab:ablation_all} demonstrate the contributions of each component in challenging scenes with complex motions and occlusions.

\begin{figure}
    \centering
    \includegraphics[width=1.0\columnwidth]{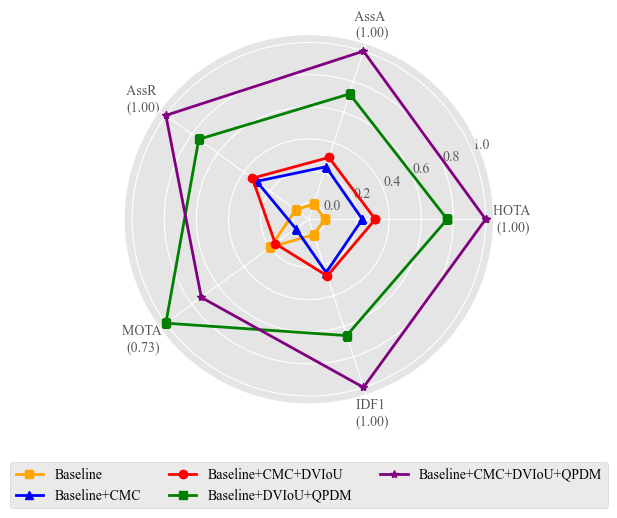}
    \caption{ Radar chart of the gains obtained through different combinations of modules on the validation set of DanceTrack. The values in the graph are obtained by min-max normalizing each metric in Table \ref{tab:ablation_all}.}
    \label{fig:ablation-dance}
\end{figure}

To more intuitively display the contribution of the modules, we also visualize the performance of the methods on the DanceTrack validation set, as illustrated in Fig. \ref{fig:ablation-dance}. We can see that each step from the base method to PD-SORT achieves improvements in most metrics.
It is worth noting that QPDM, as a module that directly utilizes pseudo-depth information, brings particularly obvious performance improvements, which further verifies the effectiveness of pseudo-depth in scenarios similar to DanceTrack.
    \begin{table}[t]
        \centering
         \caption{Ablation study on DanceTrack-val.}        
        \label{tab:ablation_all}
        \begin{tabular}{c c c | c c c c}        
        \hline
        \hline
        \textbf{CMC} & \textbf{DVIoU} & \textbf{QPDM} & \textbf{HOTA↑} & \textbf{AssA↑}  & \textbf{MOTA↑} & \textbf{IDF1↑}\\
        \hline
        & &                                   & 52.2 & 35.3 & 87.3 & 51.9 \\
        \checkmark & &                        & 52.9 & 36.4 & 87.2 & 52.8 \\
        \checkmark & \checkmark &             & 53.2 & 36.6 & 87.3 & 52.9 \\
         & \checkmark & \checkmark            & 54.7 & 38.5 & 87.5 & 54.2 \\
         \checkmark & \checkmark & \checkmark & 55.5 & 39.8 & 87.4 & 55.4 \\
        \hline
        \hline
        \end{tabular}
    \end{table}
\subsubsection{Impact of Pseudo-Depth Quantization}
We compare the QPDM module using quantized pseudo-depth as the matching metric with an alternative approach that directly uses the absolute difference (ABS) between continuous pseudo-depth values. As shown in Table \ref{tab:ablation_QPDM}, QPDM with six or more pseudo-depth intervals consistently outperforms ABS across metrics. This highlights the advantage of quantizing pseudo-depth into subintervals for robust similarity distance measurement.
    \begin{table}[t]
        \centering          
        \caption{Results of different psesudo-depth matching strategies on DanceTrack validation set.}
        \label{tab:ablation_QPDM}    
        \begin{tabular}{c | c c c c}
        \hline
        \hline
        \textbf{Matching Strategy} & \textbf{HOTA↑} & \textbf{AssA↑}  & \textbf{MOTA↑} & \textbf{IDF1↑}\\
        \hline
       ABS  & 54.8 & 38.8 & 87.4 & 54.6 \\
       QPDM (Interval Num=2)  & 54.6 & 38.7 & 87.4 & 55.0 \\
       QPDM (Interval Num=4)  & 54.2 & 38.0 & 87.5 & 53.7 \\
       QPDM (Interval Num=6)  & 55.3 & 39.6 & 87.4 & 55.2 \\
        QPDM (Interval Num=8)  & 55.5 & 39.8 & 87.4 & 55.4 \\
        QPDM (Interval Num=10) & 55.4 & 39.7 & 87.4 & 55.1 \\
        \hline
        \hline
        \end{tabular}  
    \end{table}
\subsubsection{Number of Pseudo-Depth Intervals in QPDM}
In Table \ref{tab:ablation_QPDM}, we investigate the influence of the subinterval number on the DanceTrack validation set. Specifically, we tested subinterval numbers from 2 to 10, with a step of 2. The performance gain from QPDM was low for small numbers of subintervals. We consider that fewer subinterval divisions result in fewer differences in depth and provide less guidance for distinguishing between targets. As the number of subintervals reached 6 to 8, most metrics reached the best results and dropped as it increased to 10. The reason is that too fine-grained subinterval divisions could cause an oversensitivity to changes in the targets' relative locations. Furthermore, the sparsity of the target distribution influences the choice of the ideal number of subinterval division. Generally, the tracker with a pseudo-depth subinterval number of 8 reached the most optimal metrics. Thus, we use 8 as our subinterval number for the experiments and the reported results on the test sets. 
\subsubsection{DVIoU or Standard IoU}
We also investigate the proper IoU strategies to be used in both rounds of associations, namely the regular association and the ORU in OC-SORT. Specifically, we test the standard 2D IoU and our proposed depth volume IoU (DVIoU) for similarity evaluations in the above associations.
The experimental results on the DanceTrack validation set are shown in Table \ref{tab:ablation_iou}. We can see that using DVIoU for both rounds of associations brings the best performance, which further demonstrates that the depth cue brings stable discrimination information and is able to robustly improve the tracking quality.
    \begin{table}[t]
        \centering
        \caption{Results of different IoU on DanceTrack validation set.}
        \label{tab:ablation_iou}
        \resizebox{\columnwidth}{!}{
        \begin{tabular}{c | c | c c c c}
        \hline
        \hline
        \textbf{Regular Association} & \textbf{ORU} & \textbf{HOTA↑} & \textbf{AssA↑}  & \textbf{MOTA↑} & \textbf{IDF1↑}\\
        \hline
        IoU  & IoU      & 55.2   & 39.4   & 87.4   & 55.0 \\
        DVIoU & IoU     & 55.4   & 39.7   & 87.4   & 55.3 \\
        IoU & DVIoU     & 55.2   & 39.4   & 87.4   & 55.1 \\
        DVIoU & DVIoU   & 55.5   & 39.8   & 87.4   & 55.4 \\
        \hline
        \hline
        \end{tabular}
        }
    \end{table}
\subsubsection{Impact of Complementary View}
We evaluate the effectiveness of the complementary view in pseudo-depth estimation by constructing a variant for comparison. Following SparseTrack, this variant estimates the pseudo-depth directly as the distance from the bottom of the target bounding box to the bottom of the image view. As shown in Table \ref{tab:complet-view}, incorporating the complementary view contributes to superior performance across multiple metrics. By improving the estimation robustness in boundary cases, the subsequent components DVIoU and QPDM based on pseudo-depth can provide more accurate guidance for target association.
    \begin{table}[t]
        \centering          
        \caption{Results of different psesudo-depth estimation methods on DanceTrack validation set.}
        \label{tab:complet-view}    
        \begin{tabular}{c | c c c c}
        \hline
        \hline
        \textbf{Estimation Method} & \textbf{HOTA↑} & \textbf{AssA↑}  & \textbf{MOTA↑} & \textbf{IDF1↑}\\
        \hline
        w/o complementary view  & 54.0 & 37.8 & 87.1 & 53.3 \\
        w/ complementary view  & 55.5 & 39.8 & 87.4 & 55.4 \\        
        \hline
        \hline
        \end{tabular}  
    \end{table}
\subsubsection{Validation of CMC on KF States and Historical Observations}
In addition, we also explored the effectiveness of using CMC correction for KF states as well as historical observations in our PD-SORT, and the results are shown in Table \ref{tab:ablation_cmc}. We can see that both applying CMC to KF states (CMC-KF) and historical observations (CMC-HISOB) individually can bring benefit to the tracking performance. Further, the joint application of CMC on both the KF states and historical observations brings even better overall performance.
    \begin{table}[t]
        \centering  
        \caption{Evaluation of different CMC strategies on DanceTrack validation set.}
        \label{tab:ablation_cmc}
        \begin{tabular}{c | c | c c c c}
        \hline
        \hline
        \textbf{CMC-KF} & \textbf{CMC-HISOB} & \textbf{HOTA↑} & \textbf{AssA↑}  & \textbf{MOTA↑} & \textbf{IDF1↑}\\
        \hline
          &                       & 54.7   & 38.5   & 87.5   & 54.2 \\
        \checkmark &              & 55.2   & 39.7   & 87.5   & 55.1 \\
         & \checkmark             & 54.8   & 39.4   & 87.5   & 54.4 \\
        \checkmark & \checkmark   & 55.5   & 39.8   & 87.4   & 55.4 \\
        \hline
        \hline
        \end{tabular}      
    \end{table}
\subsection{Visualization}
\begin{figure*}
    \centering
    \includegraphics[width=0.92\textwidth]{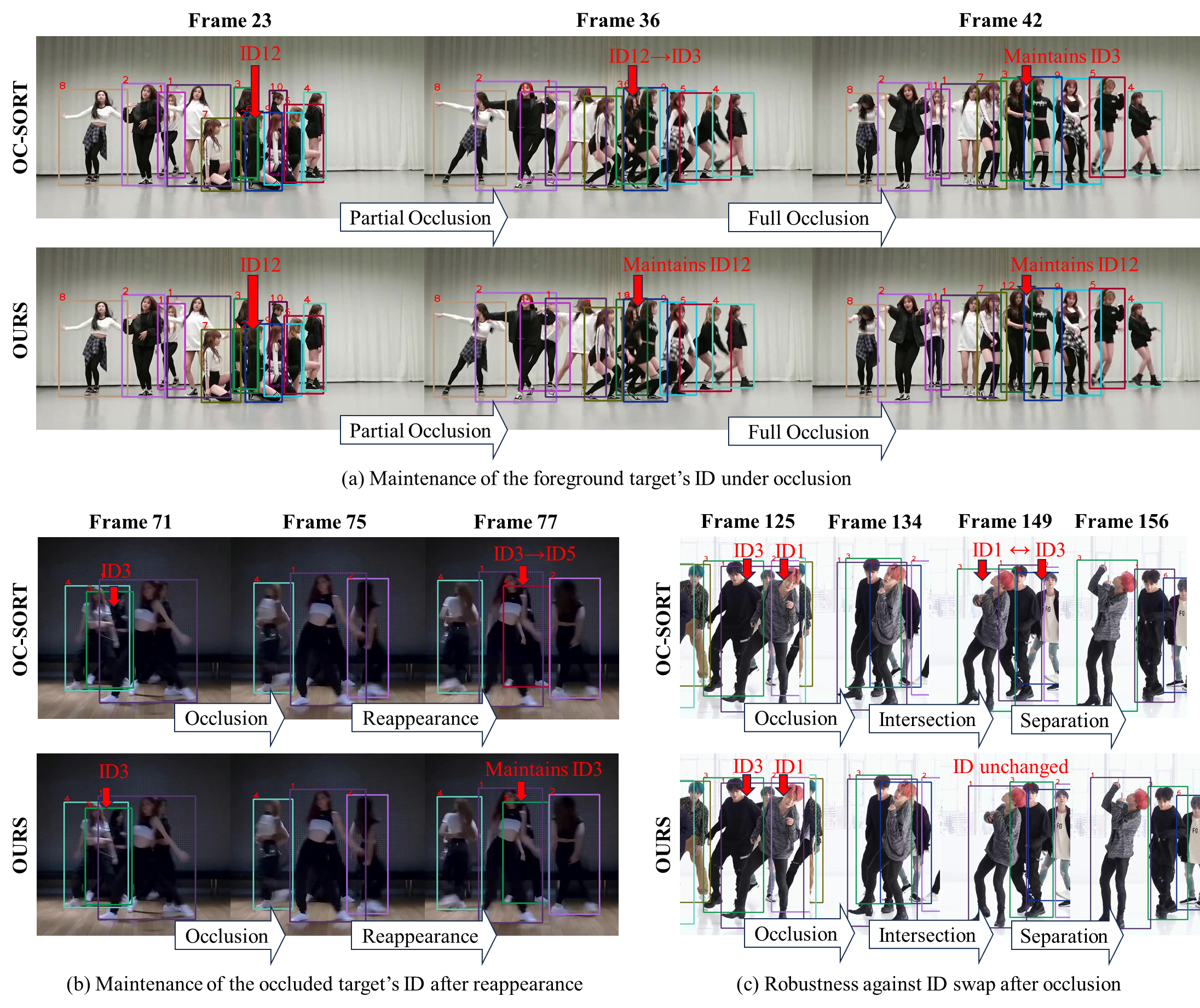}
    \caption{ Visualization of the tracking results between the 2D tracker OC-SORT and the proposed PD-SORT tracker utilizing pseudo-depth on the DanceTrack dataset. Different colors represent different identities. Our PD-SORT produces fewer identity-related association errors under occlusions. }
    \label{fig:vis-all}
\end{figure*}

The performance comparisons between the classical 2D tracker (OC-SORT) and our proposed approach (PD-SORT) utilizing pseudo-depth on DanceTrack are shown in Fig. \ref{fig:vis-all}. From the visualized results, our method can handle identity consistency problems well in challenging scenes with occlusions and nonlinear object motions, thus leading to a robust association. Specifically, PD-SORT can handle three typical kinds of occlusion-induced ID problems, namely the ID replacement of the foreground object by the occluded object, the ID reinitialization of the occluded object after reappearance, and the ID swap of objects under occlusion and trajectory intersection. In such cases, the depth of the object provides discriminative information that fixes the association failure of pure 2D information.
\subsection{Limitations}
Our experiments reveal several limitations of PD-SORT. One concern is its association ability against long-term occlusion. In such cases, if the occluded object is in quick motion, the motion consistency of the object can fail to match the reappeared object's previous trajectory. This is a common problem with motion-based MOT trackers. To solve such problems, incorporating appearance models or using learnable association matchers can be effective. 
Another concern is that our projection-based pseudo-depth estimation is performed at the instance level, without generating a full depth map of the entire image. This limits the full use of depth information. 
Besides, in highly crowded environments, the presence of numerous targets with similar pseudo-depth values and significant overlap between objects can reduce the discriminative power of pseudo-depth cues. Similarly, rapid motion changes will challenge the tracker’s ability to maintain accurate pseudo-depth estimates, potentially affecting association precision. Exploring network-based depth estimators and incorporating context-aware techniques could be potential solutions for the above issues.
In addition, although our method performs well on the HOTA metric, the performance gain on the MOTA metric is not significant and even has a slightly lower MOTA than the baseline on the MOT20 test set. This may be due to the missing of low-confidence detection results, which may be solved using an adaptive detection threshold strategy. Future work is needed to incorporate appearance cues and develop more comprehensive strategies to exploit all possible targets.

\section{Conclusion}\label{sec: conclusion}
In this paper, we demonstrate the feasibility of incorporating pseudo-depth into the object motion model in motion-based MOT. The pseudo-depth information can provide guidance for associations when 2D information fails. Consequently, we present PD-SORT, which leverages pseudo-depth to enhance the tracker's association performance. Specifically, we integrate pseudo-depth into KF and employ two simple designs, DVIoU and QPDM, to leverage the depth information in matching. Moreover, we use the camera motion compensation technique to address the camera motion. Notably, PD-SORT maintains a simple, online, real-time, and pure motion-based tracker while having better robustness against occlusions. Experiments on diverse datasets show that PD-SORT consistently outperforms its baseline and most state-of-the-art methods on scenes with different motions and densities. The performance gain is especially significant in dense scenes with similar appearances and nonlinear object motions. 
Specifically, PD-SORT achieves 58.2 HOTA, 80.6 DetA, 42.1 AssA, and 57.5 IDF1 on the DanceTrack test set with 28.7 FPS, which is +3.6 HOTA, +0.2 DetA, +1.9 AssA, and +2.9 IDF1 over the baseline.

In future work, we plan to explore more effective depth utilization strategies and integrate learnable association modules to further enhance tracking performance. Also, we plan to incorporate additional context-aware information (e.g., actual depth data, appearance cues, infrared data) to improve the tracker's robustness in complex scenes that contain highly crowded and fast-moving objects. Finally, we hope the occlusion-robust characteristic and generalization ability of PD-SORT can make it attractive for application in consumer electronics and inspire future research to further investigate the depth cues and make MOT methods more practical.


{
    \bibliographystyle{IEEEtran}
    \bibliography{my_references}
}

\end{document}